\documentclass[conference]{IEEEtran}
\IEEEoverridecommandlockouts
\usepackage{cite}
\usepackage{amsmath,amssymb,amsfonts}
\usepackage{algorithmic}
\usepackage{graphicx}
\usepackage{textcomp}
\usepackage{xcolor}
\usepackage{subcaption}
\usepackage[inkscapelatex=false]{svg}

\usepackage{graphicx}%
\usepackage{multirow}%
\usepackage{amsthm}%
\usepackage{mathrsfs}%
\usepackage{manyfoot}%
\usepackage{booktabs}%
\usepackage{listings}%
\usepackage{url}

\newcommand{\EX}{{\mathbb E}}
\newcommand{\R}{{\mathbb R}}

\newcommand{\D}{{\mathcal{D}}}

\newtheorem{theorem}{Theorem}
\newtheorem{proposition}[theorem]{Proposition}%

\def\BibTeX{{\rm B\kern-.05em{\sc i\kern-.025em b}\kern-.08em
    T\kern-.1667em\lower.7ex\hbox{E}\kern-.125emX}}
\begin{document}

\title{Harnessing Contrastive Learning and Neural Transformation for Time Series Anomaly Detection\\
}

\author{\IEEEauthorblockN{Jiazhen Chen}
\IEEEauthorblockA{\textit{dept. Statistics and Actuarial Science} \\
\textit{University of Waterloo}\\
Waterloo, Canada \\
j385chen@uwaterloo.ca}
\and
\IEEEauthorblockN{Mingbin Feng}
\IEEEauthorblockA{\textit{dept. Statistics and Actuarial Science} \\
\textit{University of Waterloo}\\
Waterloo, Canada \\
ben.feng@uwaterloo.ca}
\and
\IEEEauthorblockN{Tony S. Wirjanto}
\IEEEauthorblockA{\textit{dept. Statistics and Actuarial Science} \\
\textit{University of Waterloo}\\
Waterloo, Canada \\
twirjanto@uwaterloo.ca}
}

\maketitle

\begin{abstract}

Time series anomaly detection (TSAD) plays a vital role in many industrial applications. While contrastive learning has gained momentum in the time series domain for its prowess in extracting meaningful representations from unlabeled data, its straightforward application to anomaly detection is not without hurdles. Firstly, contrastive learning typically requires negative sampling to avoid the representation collapse issue, where the encoder converges to a constant solution. However, drawing from the same dataset for dissimilar samples is ill-suited for TSAD as most samples are ``normal'' in the training dataset.  Secondly, conventional contrastive learning focuses on instance discrimination, which may overlook anomalies that are detectable when compared to their temporal context. In this study, we propose a novel approach, CNT, that incorporates a window-based contrastive learning strategy fortified with learnable transformations. This dual configuration focuses on capturing temporal anomalies in local regions while simultaneously mitigating the representation collapse issue. Our theoretical analysis validates the effectiveness of CNT in circumventing constant encoder solutions. Through extensive experiments on diverse real-world industrial datasets, we show the superiority of our framework by outperforming various baselines and model variants.

\end{abstract}

\begin{IEEEkeywords}
time series anomaly detection, contrastive learning, unsupervised learning, neural networks
\end{IEEEkeywords}
\section{Introduction}
Modern industrial systems rely on IoT technology for communication and information exchange among sub-systems. Due to the vast scale of these distributed systems, unanticipated events such as cyber-attacks can cause system downtime and may even lead to catastrophic failures~\cite{mohammadi2018deep,gdn,mad-gan}. For instance, in smart power grids, malicious activities can disrupt electricity distribution, leading to widespread blackouts and significant economic losses. Therefore, an efficient and accurate time series anomaly detection (TSAD) system is crucial for monitoring critical infrastructure and providing timely anomaly alerts, helping prevent failures and safeguard data. Despite the importance of TSAD, its efficacy is often constrained by the rarity of anomalies and the resource-intensive nature of obtaining accurate labels. These issues limit the practicality of supervised methods and drive the need for unsupervised TSAD approaches. Deep learning models excel in this regard, as they can learn complex feature representations without reliance on manual labeling.

Deep learning-based unsupervised TSAD primarily focuses on identifying patterns of normality within unlabeled data, with the expectation that anomalies will manifest as deviations from these identified patterns. Existing methods primarily include forecasting-based~\cite{lstm-ndt,tcnanomaly,gdn,fusagnet,gta,chen2023multivariate} and reconstruction-based approaches~\cite{mad-gan,usad, mscred,tranAD,anomalyTransformer,chen2023adversarial}. The former assumes normal instances are predictable, but their performances can falter when temporal patterns are unpredictable or rapidly changing~\cite{lstmed}. The latter, despite not depending on sample extrapolation, can run into overfitting issues due to an undue emphasis on pixel-wise reconstructions. In recent years, contrastive learning has emerged as a significant area of interest in various domains~\cite{he2020momentum,chen2020simple,saeed2021contrastive,shi2019deepchannel}, lauded for its capability to distill valuable feature representations from data. Unlike traditional approaches, contrastive learning operates on the latent feature space, advocating for the alignment of similar instances and the segregation of dissimilar ones. In recent years, a growing number of studies have applied contrastive learning to anomaly detection and shown promising results in various domains such as computer vision, natural language processing, and speech processing~\cite{geoTransformation,robustAndUncertainty,inlier,hojjati2022self,qi2023logencoder}.

Yet applying contrastive learning to TSAD presents unique challenges. First, anomalies in time series data are context-dependent, necessitating an analysis that not only considers each instance but also understands its relationship to historical patterns. Secondly, contrastive learning typically relies on selecting negative samples to avoid representation collapse, where models converge to trivial solutions mapping different inputs to indistinguishable representations. However, deriving negative samples from a pool of predominantly normal instances is ill-suited for anomaly detection. moreover, the generation of ``similar'' positive samples through data augmentation, such as geometric transformations for images (e.g., cropping and rotating)~\cite{geoTransformation}, is not directly applicable to time series data due to their intricate temporal dependencies.

In this work, we propose a simple but effective TSAD method, CNT, to address the aforementioned challenges. Motivated by Contrastive Predictive Coding (CPC)~\cite{cpc}, we implement a relatively simple but effective temporal-wise contrasting where each instance is defined by a prediction time window and is pulled closer to a corresponding context window through a contextual contrastive loss. This strategy naturally discerns anomalies by highlighting significant divergences from historical patterns. However, this alignment alone can result in the representation collapse issue: the feature encoders of both windows produce constant representations irrespective of their input. In traditional contrastive learning, this issue is often circumvented by using negative samples, ensuring diversity in the learned representations. To avoid incorrectly selecting negative samples within the same normal class,  we enhance the temporal-wise contrasting process by applying learnable neural transformations to the latent space of the prediction window, ensuring that the resulting representations remain diverse yet consistent with the original inputs. Subsequently, temporal-wise contrasting is executed between these transformed windows and their past observations, with anomalies anticipated to disrupt this contextual consistency more than normal instances. Lastly, our theoretical analysis indicates that this duel-loss circumvents the issue of representation collapse without the need for negative samples. Extensive experiments are conducted on five real-world industrial datasets to demonstrate the superior performance of our proposed approach over existing baseline methods.

\section{Methodology}
\subsection{Problem Formulation}
\label{sec: problem_formulation}
\begin{figure}[t!]
    \centering
\includegraphics[width=1\linewidth]{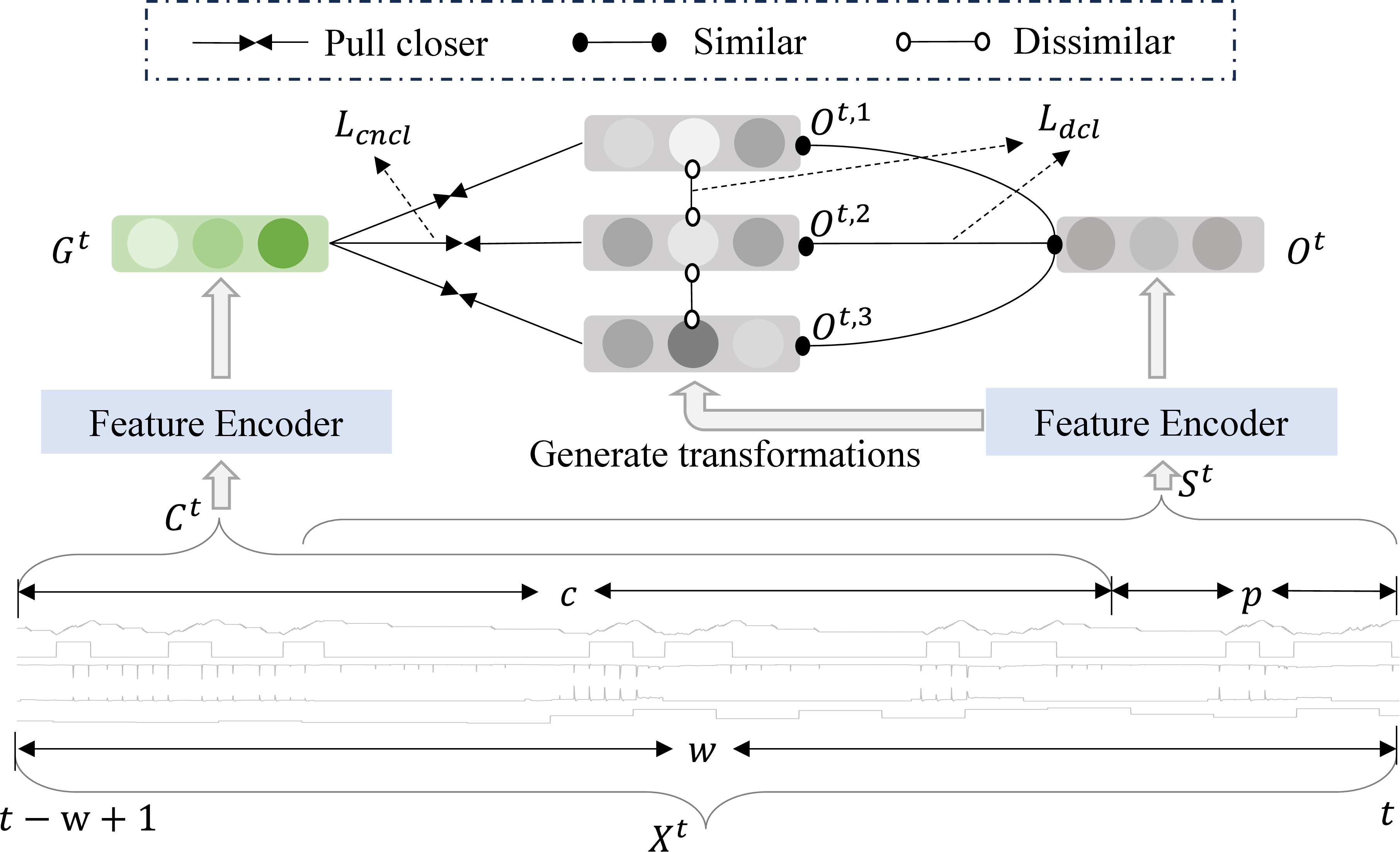} 
\caption{An illustration of the CNT framework.}
\label{fig: framework}
\end{figure}

We are given a multivariate time series $X \in \R^{N\times T}$, collected over
$T$ time ticks with $N$ features, as the training set. The goal is to detect anomalies in a test set, which contains a time series with $N$ features over a different time span. The training set $\D$ consists of sub-sequences extracted from $X$, i.e., $X$ is converted into sliding windows with stride 1. Here, a sample $X^t$ represents a collection of time ticks within a sliding
window of length $w$: $X^t=\{X_{\cdot i}\}_{i=t-w+1}^t$, where $X_{\cdot i} \in \R^{N}$ is the multivariate time series value at timestamp $i$. Our objective is to assign an anomaly score to each time tick in the test set, which is later thresholded to classify each tick as normal (0) or abnormal (1). Furthermore, in our framework, each sample $X^t$ is segmented into two overlapping sub-sequences of length $c$: a context sequence $C^t$ and a prediction sequence $S^t$, where $S^t$ is shifted forward from $C^t$ by a small time window of length $p=w-c$, i.e., $C^t=\{X_{\cdot i}\}_{i=t-w+1}^{t-p}$ and $S^t = \{X_{\cdot i}\}_{i=t-c}^{t}$. The time window from $t-p+1$ to $t$ is referred to as the prediction window, while $t-w+1$ to $t-p$ is the context window. A visualization of notations is shown in Figure~\ref{fig: framework}.

\subsection{Overview}
This section presents an overview of the proposed framework, CNT, which includes a feature encoder module and a neural transformation module
with both jointly optimized using two contrastive losses (See Figure~\ref{fig: framework} for an illustration). Specifically, the framework begins with the feature encoder, which converts the prediction and context
sequence of a given sample into hidden representations. Next, the neural transformation network transforms the latent representation of the prediction sequence into multiple latent representations. Finally, model parameters are optimized under two loss functions: a contextual neural contrastive loss and a discriminative contrastive loss. 
The former aims at minimizing the distance between the context sequence and the transformations of prediction sequence. The latter ensures that the transformed sequences remain diverse while maintaining the original semantics of the prediction sequence.



\subsection{Framework}
\label{sec: optimizatino}

\subsubsection{Feature Encoder}

\label{sec: feature encoder}

The feature encoder aims at capturing temporal dependencies within a prediction or context sequence ($C^t$ or $S^t$), thereby facilitating the detection of anomalies in the prediction window by identifying anomalous temporal patterns with respect to the context window. 
Following~\cite{wu2020connecting}, we use a temporal convolutional network (TCN) as the base feature encoder, with a dilated inception layer (DIL)~\cite{wu2020connecting} replacing the temporal convolution layer to support different kernel sizes. The encoded representations of $S^t$ and $C^t$ are denoted as $O^t$ and $G^t$, respectively.

\subsubsection{Contextual Contrastive Loss}
Our framework assumes that normal sequences change mildly compared to abnormal sequences. To implement this idea, we minimize the distance between the hidden representations of the prediction and context sequences for normal samples. Specifically, let $g(\cdot; W): \R^{N \times c} \rightarrow \R^d$ be the feature encoder with learnable weights $W$. We define a contextual contrastive loss as:
\begin{align}
   & L_{\text{ccl}}:= \EX_{X^t \sim \D}{ \|g(S^t;W) - g(C^t;W)\|}^2_2
    \label{eq. raw contextual one-class objective}
\end{align}
However, minimizing this loss alone can lead to representation collapse, where the encoder outputs constant values. To address this, inspired by NeuTralAD~\cite{neuralAD}, we combine the contextual contrastive loss with neural transformation learning, ensuring distinct outputs across different contexts.



\subsubsection{Combining With Neural Transformations}

Neural transformation learning learns diverse transformations that maintain the original sample's semantic information. We leverage this approach to prevent representation collapse by contrasting the multifaceted transformed representations of the prediction sequence $O^t$ against the context sequence $G^t$.

Given $O^t$, we project it into $K$ latent representations $\{O^{t,i}\}_{i=1}^K$ using distinct transformation networks $T=\{T_1, ..., T_K|T_k:\R^d \rightarrow \R^d\}$. Each $T_k$ is a three-layer MLP with hidden dimension $d$. Unlike the original approach in~\cite{neuralAD}, we apply transformations directly in the latent space instead of raw inputs. Thereby, it reduces computational complexity by avoiding repeated passes through the feature encoder.


To maintain diversity and semantic consistency, a discriminative contrastive loss is applied on $O^t$ and $\{O^{t,i}\}_{i=1}^K$:
\begin{equation}
\resizebox{.9\hsize}{!}{$
    L_{\text{dcl}}:= -\EX_{X^t \sim \D} \left[ \sum_{k=1}^K \log{\frac{h(O^t,O^{t, k})}{h(O^t,O^{t, k})+\sum_{l \neq k} h(O^{t, k}, O^{t, l})}} \right] 
    $}
\end{equation}
where $h(x, y) = \exp(\frac{x^Ty}{{\tau\|x\|}_2{\|y\|}_2})$. $\tau$ is a hyper-parameter called temperature. We then extend $L_{\text{ccl}}$ to a contextual neural contrastive loss $L_{\text{cncl}}$, which is defined as:
\begin{align}
        & L_{\text{cncl}} := \EX_{X^t \sim \D} \left[ \sum_{k=1}^K { \|O^{t, k} - G^t\|}^2_2 \right]
        \label{eq. contextual neural contrastive loss}
\end{align}
Here, the transformed representations of the prediction sequence are pulled closer to those of the context sequence.

Finally, the parameters of the feature encoder and neural transformations are optimized under a contextual discriminative contrastive loss $L_{\text{cdcl}}$, defined as:
\begin{align}
    & L_{\text{cdcl}} := L_{\text{cncl}} + L_{\text{dcl}} 
    \label{eq. contextual discriminative contrastive loss}
\end{align}

During inference, the loss in Eq.~\eqref{eq. contextual discriminative contrastive loss} for an input at time $t+p-1$, i.e., $X^{t+p-1}$, is used as the anomaly score for time $t$. Thus, a shorter prediction window size $p$ can introduce a smaller detection delay.

\subsection{Theoretical Analysis}
In this section, we discuss the representation collapse issue that arises when applying solely the contextual contrastive loss and demonstrate how integrating a discriminative contrastive loss mitigates this problem.  The contextual contrastive loss in Eq.~\eqref{eq. raw contextual one-class objective} can lead to a trivial solution where the feature encoder outputs a constant value, e.g., setting all weights to zero. However, constant encoders are not optimal for minimizing the contextual discriminative contrastive loss in Eq.~\eqref{eq. contextual discriminative contrastive loss}, as shown in the following proposition.

\label{sec:theoretical analysis}
\begin{proposition}
If there exists a choice of parameters for the feature encoder $g(\cdot)$ and the $K$ transformation networks $\{T_k(g(\cdot))\}_{k=1}^K$ such that $L_\text{dcl} < K \log K$, then the constant parameter setting $T_k(g(\cdot)) = g(\cdot) = c, \forall c \in \R^d$ is not an optimal solution for minimizing $L_{\text{cdcl}}$.
\label{props: non-constant encoder}
\end{proposition}

\textbf{Proof.} For simplicity, let $g_k(\cdot)$ denote $T_k(g(\cdot))$. First, note that when $g_k(\cdot) = g(\cdot) = c$, the contextual discriminative contrastive loss $L_{\text{cdcl}}$ simplifies to $K \log K$. Specifically:
\begin{align}
L_{\text{cdcl}} 
& = K \| c - c \|^2_2-K\log {\frac{h(c, c)}{h(c, c)+(K-1)h(c, c)}} \\
& = K\log K
\end{align}
Assuming that there exists a parameter setting such that $L_{\text{dcl}} < K \log K$, let $\epsilon = K \log K - L_{\text{cdcl}}$. Since the scoring function $h(\cdot, \cdot)$ in $L_{\text{dcl}}$ is defined as $h(g_k(\cdot), g(\cdot)) = \exp(\frac{g_k(\cdot)^Tg(\cdot)}{\tau\|g_k(\cdot)\|_2\|g(\cdot)\|_2})$, we can rescale the functions $\{g_k(\cdot)\}_{k=1}^{K}$ and $g(\cdot)$ during training such that $\| g_k(x)\|_2 \leq \frac{\epsilon}{2K}$ and $\| g(x)\|_2 \leq \frac{\epsilon}{2K}$ for all $x$. This rescaling is possible while ensuring that $h(g_k(x), g(x))$ remains the same, since cosine similarity between two vectors is invariant to their norms. Hence, we have:
\begin{align}
L_{\text{cncl}} = & \EX_{X^t\sim \D} \left[ \sum_{k=1}^K \| g_k(O^t) - g(G^t) \|^2_2 \right] \\
\leq & \EX_{X^t\sim \D}\left[  \sum_{k=1}^K (\| g_k(O^t)\|^2_2 + \| g(G^t) \|^2_2) \right]\\
< & \sum_{k=1}^K (\frac{\epsilon}{2K}+\frac{\epsilon}{2K}) =  \epsilon
\end{align}
Therefore, we have $L_{\text{cncl}} + L_{\text{dcl}} < \epsilon + L_{\text{dcl}} = \epsilon + K \log K - \epsilon = K \log K$. \qed

The assumption in Proposition~\ref{props: non-constant encoder} is empirically validated with $\mathcal{L}_{\text{dcl}} \ll K\log K$ after training (See in Section~\ref{sec: learning situation}). 





\section{Experimental Results}
\label{s:Experimental}
\subsection{Experimental Setting}
\subsubsection{Datasets}
We evaluate our framework using five real-world industrial TSAD datasets: 
 SWaT~\cite{mathur2016swat}, WADI~\cite{ahmed2017wadi}, HAI~\cite{hai21.03}, SMAP and MSL~\cite{lstm-ndt}. Each dataset has a unlabeled training set, and a separate test set with anomalies. Following~\cite{gdn}, we down-sampled the SWaT, WADI, and HAI datasets to 10-second intervals using the median values, assigning anomaly labels (0 or 1) based on the mode.

\subsubsection{Baseline Methods and Evaluation Metrics}
We compare our approach to several recent anomaly detection methods, including MAD-GAN~\cite{mad-gan}, USAD~\cite{usad}, MSCRED~\cite{mscred}, TranAD~\cite{tranAD}, NeuTralAD~\cite{neuralAD},  AnomalyTransformer~\cite{anomalyTransformer}, COCA~\cite{coca},  DCdetector~\cite{yang2023dcdetector} and CST-GL~\cite{zheng2023correlationaware}. 
In line with previous studies~\cite{omnianomaly}~\cite{anomalyTransformer}~\cite{tranAD}~\cite{gdn}, 
we use the best point-wise F1 (denoted as F1$^p$) and best composite F1 (denoted as F1$^{c}$)~\cite{garg2021evaluation} score as the performance metrics. 

\subsubsection{Implementation}
We train the model with the Adam optimizer~\cite{kingma2014adam} at a 0.001 learning rate for 30 epochs. 20\% of the training set serves as a validation set, with the best epoch selected based on validation loss. The hidden dimension $d$ is set to 64, with a maximum of 8 TCN blocks. The temperature $\tau$ and the number of transformations $K$ for the DCL loss are 0.1 and 6, respectively. The sequence length is 30, and the prediction window size $w$ is 5. Performance results are averaged over five independent runs.



\subsection{Comparison Study}  Table~\ref{tab:comparison_result_f1} show that CNT consistently outperforms all baseline methods across datasets, exceeding the best-performing baseline by over 10\% in both point-wise F1 and composite F1. Unlike forecasting-based and reconstruction-based methods (USAD, CST-GL, AnomalyTransformer, etc.) that focus on pixel-level details, CNT emphasizes contrasting latent representations, making it more robust and less prone to overfitting. Compared to the other contrastive learning-based method (COCA, DCdetector), CNT employs more diverse augmentations at the latent level, leading to a deeper understanding of the data without relying on manual time series augmentations. Additionally, CNT surpasses NeutralAD by a notable margin. NeuTralAD focuses on the entire time window, which can dilute anomaly detection when normal trends dominate. In contrast, CNT emphasizes contextual semantic similarity, allowing it to detect subtle anomalies by identifying deviations in shorter prediction window relative to its historical context.

\begin{table}[t!]
    \centering
        \caption{Performance comparison based on point-wise F1 (F1$^{p}$) and composite F1 (F1$^{c}$), with the best and second-best methods highlighted in bold and underline, respectively. }
       \resizebox{\linewidth}{!}{
    \begin{tabular}{cccccccccccc}\\ \toprule
               &   \multicolumn{2}{c}{\textbf{SWaT}}  & \multicolumn{2}{c}{\textbf{WADI}} & \multicolumn{2}{c}{\textbf{MSL}}& \multicolumn{2}{c}{\textbf{SMAP}}& \multicolumn{2}{c}{\textbf{HAI}} \\ \midrule
               \textbf{Methods} & \textbf{F1}$^{p}$  & \textbf{F1}$^{c}$ & \textbf{F1}$^{p}$  & \textbf{F1}$^{c}$ & \textbf{F1}$^{p}$  & \textbf{F1}$^{c}$  & \textbf{F1}$^{p}$  & \textbf{F1}$^{c}$& \textbf{F1}$^{p}$  & \textbf{F1}$^{c}$   \\ \midrule

USAD & 76.8 & 45.4 & 28.2 & 42.0 & 23.6 & 42.0 & 28.1 & 31.7 & 39.7 & \textbf{70.8} \\MAD-GAN & 73.6 & 44.4 & 15.9 & 36.6 & 20.2 & 28.4 & 24.3 & 28.6 & 5.7 & 7.2 \\MSCRED & 75.9 & 37.1 & 29.6 & 36.4 & 27.9 & 41.7 & 23.0 & 23.6 & 36.9 & 58.0 \\NeuTralAD & 76.1 & \textbf{75.0} & 30.4 & 48.6 & 21.0 & 22.3 & 23.2 & 23.3 & 32.0 & 37.4 \\TranAD & \underline{76.9} & 46.4 & 27.3 & 43.1 & 24.6 & 42.3 & 23.5 & 26.3 & 43.5 & \underline{69.9} \\AnomalyTrans & 12.9 & 32.2 & 5.0 & 10.4 & 8.9 & 27.1 & 8.3 & 24.5 & 5.3 & 17.2 \\COCA & 67.9 & 58.8 & 20.6 & 34.3 & 22.9 & 36.8 & 24.9 & 26.5 & 32.3 & 57.1 \\DCdetector & 21.9 & 22.7 & 11.0 & 15.7 & 19.3 & 25.6 & 24.2 & 26.2 & 3.9 & 4.5 \\CST-GL & 76.8 & 59.1 & 34.1 & \underline{58.2} & 22.8 & 24.4 & 24.7 & 31.3 & 14.6 & 18.4 \\
\midrule

CCL & 27.7 & 43.9 & \underline{35.4} & 57.5 & 28.2 & 55.0 & 26.5 & 51.1 & \underline{48.9} & 56.3 \\DCL & 74.4 & 61.4 & 26.9 & 36.6 & 20.9 & 42.4 & \underline{34.1} & \underline{53.0} & 38.2 & 57.8 \\ CCL+Neg & 22.9 & 34.7 & 10.8 & 38.7 & 18.8 & 42.5 & 22.9 & 30.8 & 15.4 & 39.9 \\ CCL+Vicreg  & 22.7 & 48.0 & 29.0 & 42.4 & \textbf{38.8} & \underline{60.5} & 24.5 & 51.2 & 36.3 & 42.7 \\ \midrule
CNT (ours) & \textbf{77.1} & \underline{72.3} & \textbf{36.7} & \textbf{64.6} & \underline{28.9} & \textbf{60.9} & \textbf{34.5} & \textbf{57.4} & \textbf{50.0} & 65.1 \\

   \bottomrule
    \end{tabular}}
    \label{tab:comparison_result_f1}
\end{table}

\subsection{Ablation Study}
We further compare the performance of CNT with several model variants (bottom of Table~\ref{tab:comparison_result_f1}). The CCL setting or DCL setting trains the feature encoder using only $L_{ccl}$ or $L_{dcl}$, respectively.
 To explore alternative ways of preventing representation collapse without neural transformation learning, we test two additional variants: CCL + Neg, inspired by\cite{cpc}, introduces negative samples into the CCL framework by randomly selecting context windows from the batch as negatives and training with an infoNCE loss using $l_2$ distance. CCL + Vicreg, based on~\cite{vicreg}, adds a regularization term to the CCL loss to ensure that the latent representations have non-zero variance, thus preventing constant encoder outputs. 
 Overall, CNT outperforms all these variants on average across multiple datasets, which further validates the design choices in our model.

\subsection{Analysis Of Training Losses}
\label{sec: learning situation}
\begin{figure}[t!]
\begin{subfigure}[b]{0.49\linewidth}
    \centering
\includegraphics[height=2.4cm]{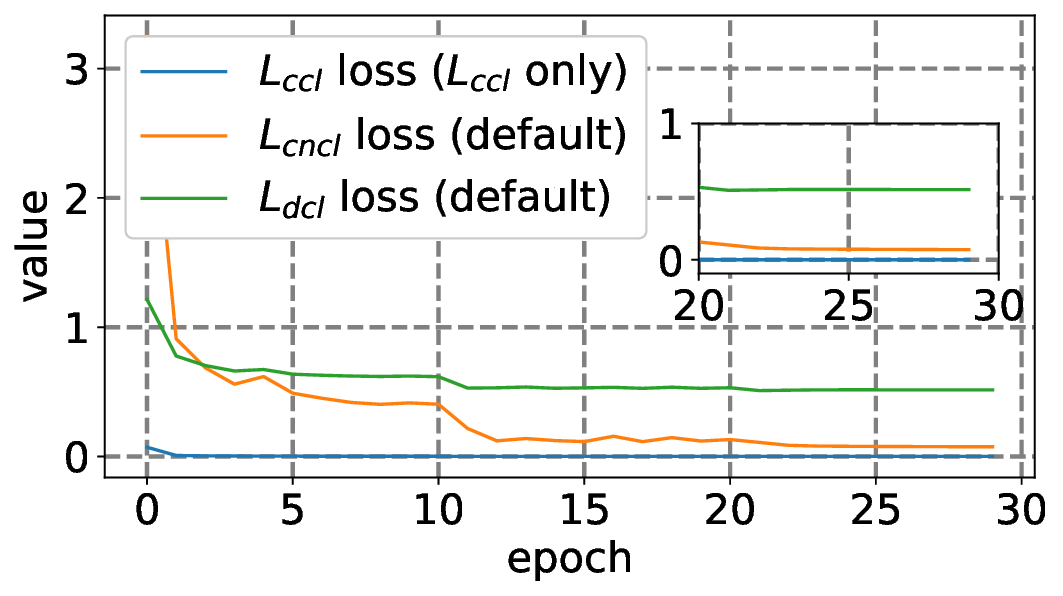}  
\caption{Loss vs. epoch (Wadi)}
\end{subfigure}
\begin{subfigure}[b]{0.49\linewidth}
    \centering
\includegraphics[height=2.4cm]{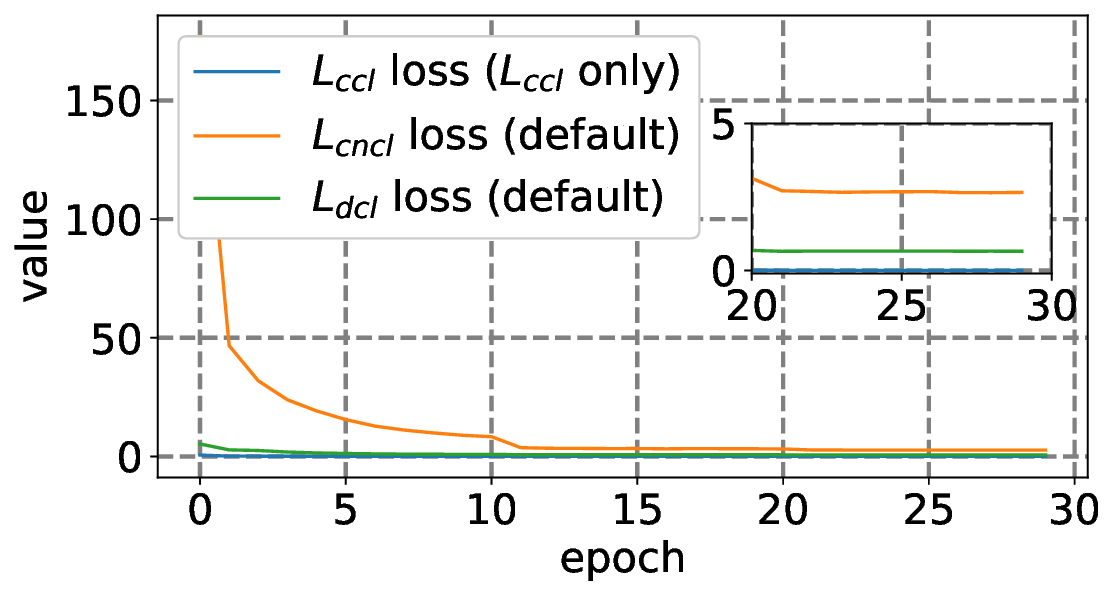}  
\caption{Loss vs. epoch (SWaT)}

\end{subfigure}

\caption{Loss value per epoch on Wadi and SWaT.}
\label{fig: loss_decay}
\end{figure}

Figure~\ref{fig: loss_decay} shows the loss values per epoch for our model training under purely $L_{\text{ccl}}$ loss or training under both $L_{\text{cncl}}$ and $L_{\text{dcl}}$ (the default setting). It is clear that $L_{\text{ccl}}$ decays quickly to 0 without the regularization of $L_{\text{dcl}}$. In contrast, with the additional constraints of $L_{\text{dcl}}$, $L_{\text{cncl}}$ decays much slowly compared to the former. 
This implies that $L_{\text{dcl}}$ does add additional complexity to the CCL task, preventing from learning trivial feature representations. Furthermore, the loss value of $L_{\text{dcl}}$ quickly falls below $K\log K$ (4.67 when $K=6$, the default parameter setting used in our experiments). This verifies the assumption in Proposition~\ref{props: non-constant encoder} is empirically applicable. 

\subsection{Data Visualization}
\begin{figure}[t!]
    \centering
        \includegraphics[height=5.05cm]{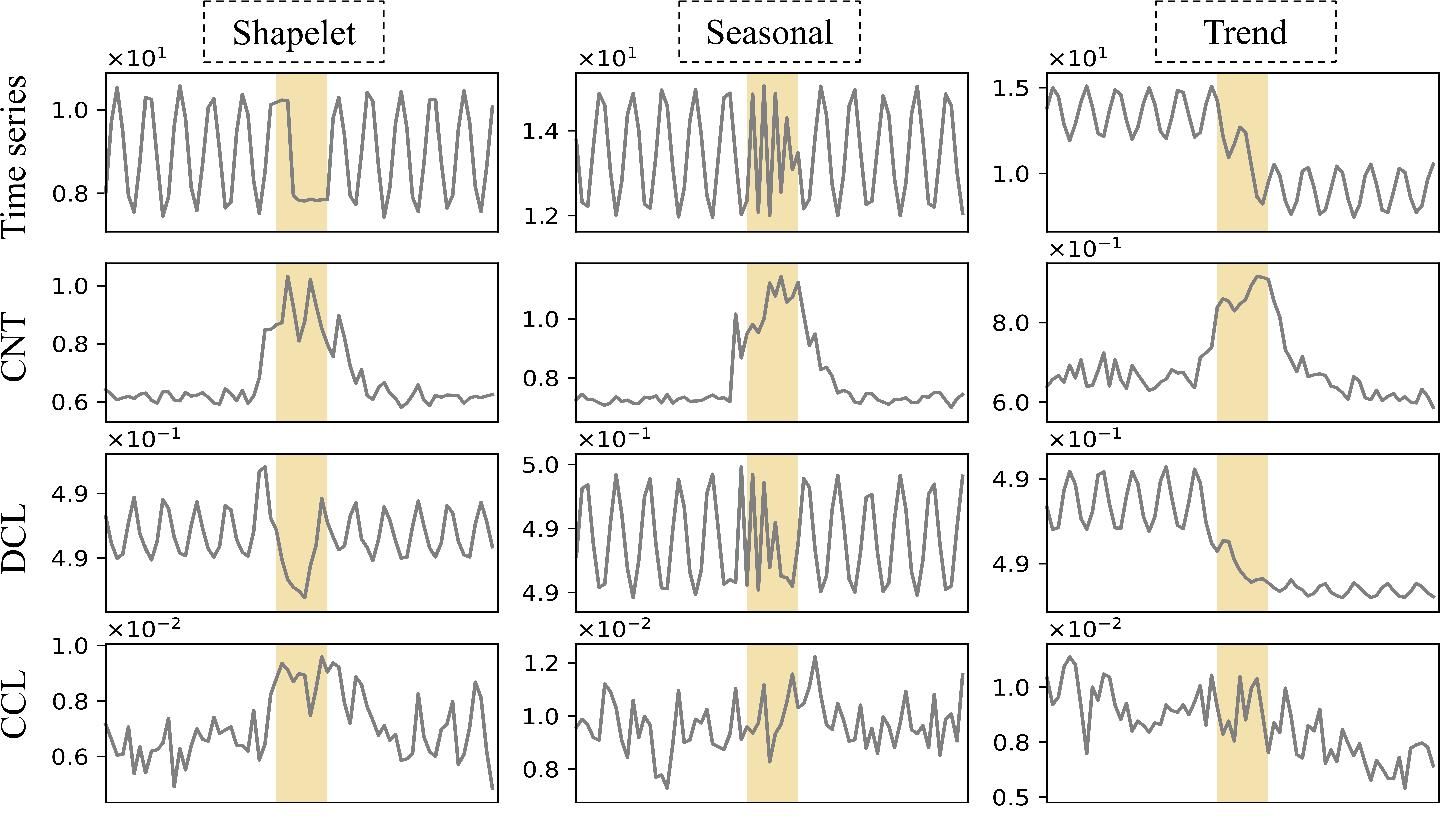}
        \caption{The top row shows the original time series, while subsequent rows display the anomaly scores from CNT, DCL, and CCL. Anomalous regions are highlighted in yellow.}
    \label{fig: tods showcase}
\end{figure}

To visually examine the effectiveness of our models, we conducted additional experiments on the univariate TODS dataset~\cite{tods}, chosen for its clear classification of anomaly types (e.g., shapelet, seasonal, and trend anomalies). Figure~\ref{fig: tods showcase} presents the anomaly detection scores generated by the CNT and its two direct variants. CNT consistently produces sharper anomaly scores in the anomaly regions, showing high sensitivity to abnormal patterns. In contrast, DCL struggles, particularly with trend anomalies, showing weaker scores and under-detecting subtle irregularities. CCL performs better than DCL but still fails to match CNT's precision, occasionally misidentifying normal regions as anomalies due to the representation collapse issue. Overall, CNT excels in detecting diverse anomalies, whereas DCL and CCL are limited by their broader focus and vulnerability to oversimplified representations.

\section{Conclusions}

In our study, we propose a contextually-aware contrastive learning framework for time series anomaly detection. By utilizing a temporal-wise contrasting strategy, we effectively compared latent representations of the predict and context windows, while addressing the representation collapse problem by augmenting the contextual contrastive loss with neutral transformation learning. This refinement enhanced the complexity and effectiveness of our approach, resulting in superior performance over a variety of baseline time series anomaly detection methods on real-world industrial datasets.

\clearpage

\bibliographystyle{IEEEtran}
\bibliography{reference}
\end{document}